\documentclass[review]{elsarticle}

\usepackage{times}
\usepackage{enumerate}
\usepackage{amsmath}
\usepackage{bbm}
\usepackage{amsfonts}
\usepackage{epsfig,epstopdf}
\usepackage{graphicx}
\usepackage{subfigure}
\usepackage[ruled]{algorithm}
\usepackage{algorithmic}
\usepackage{multirow}
\usepackage{flushend}
\usepackage{makecell}
\usepackage{lineno,hyperref}
\usepackage{xcolor}

\usepackage{amssymb}
\usepackage{graphicx}
\usepackage{url}
\usepackage{amsthm}
\newtheorem{theorem}{Theorem}

\modulolinenumbers[5]

\newcommand{\xucomment}[1]{}

\begin{document}\sloppy
\begin{frontmatter}
\title{Self-Paced Multi-Task Clustering}

\author[Uestcaddress]{Yazhou Ren\corref{mycorrespondingauthor}}
\cortext[mycorrespondingauthor]{Corresponding author}
\ead{yazhou.ren@uestc.edu.cn}

\author[Uestcaddress]{Xiaofan Que}
\author[mysecondaryaddress]{Dezhong Yao}
\author[Uestcaddress]{Zenglin Xu}

\address[Uestcaddress]{School of Computer Science and Engineering, University of Electronic Science and Technology of China, Chengdu, 611731, China}
\address[mysecondaryaddress]{Key Laboratory for NeuroInformation of Ministry of Education, School of Life Science and Technology, University of Electronic Science and Technology of China, Chengdu, 611731, China}


\begin{abstract}
Multi-task clustering (MTC) has attracted a lot of research attentions in machine learning due to its ability in utilizing the relationship among different tasks. Despite the success of traditional MTC models, they are either easy to stuck into local optima, or sensitive to outliers and noisy data. To alleviate these problems, we propose a novel self-paced multi-task clustering (SPMTC) paradigm. In detail, SPMTC progressively selects data examples to train a series of MTC models with increasing complexity, thus highly decreases the risk of trapping into poor local optima. Furthermore, to reduce the negative influence of outliers and noisy data, we design a soft version of SPMTC to further improve the clustering performance. The corresponding SPMTC framework can be easily solved by an alternating optimization method. The proposed model is guaranteed to converge and experiments on real data sets have demonstrated its promising results compared with state-of-the-art multi-task clustering methods.
\end{abstract}


\end{frontmatter}

\section{Introduction}
\label{sec:Intro}
Clustering is an unsupervised classification problem that groups samples into different clusters. 
Among different clusters, examples have no overlap, while they are similar to each other in the same cluster \cite{jain1999data,Ren2018Big}. Clustering can not only be used to find out the inner structure of data but also be used as the precursor of other machine learning tasks \cite{xu2005survey}. 
Over the past decades, many clustering approaches have been developed, such as partitional algorithms (e.g., $k$-means type clustering \cite{MacQueen:some,HUANG2018Anew}), hierarchical clustering \cite{jain1999data}, density-based clustering \cite{ester1996density,Ren2018SSDC}, distribution-based clustering (e.g., Gaussian mixture model \cite{Bishop2006Pattern_SR}), clustering based on non-negative matrix factorization (NMF) \cite{Lee01algorithmsfor,Huang2018neucom,HUANG2018Self}, 
mean shift clustering \cite{Dorin2002mean,Ren2014weighted,Ren2014Boosted}, ensemble clustering \cite{Strehl:cluster,Ren2013WOEC,Ren2017WOEC}, etc.
All of the abovementioned methods can only tackle single task and are considered as single-task clustering.

Considering the relationship among similar tasks, multi-task learning is proposed to learn the shared information among tasks \cite{liu2009multi} and it is proved to achieve great performance in many applications such as medical diagnosis 
\cite{bickel2008multi,qi2010semi}
, web search ranking \cite{chapelle2010multi}, computer vision \cite{li2014heterogeneous}, and so on.

Tasks being closely related to each other is a common issue in real clustering problems. Consequently, multi-task clustering (MTC) that uses the information shared by multiple related tasks is proposed \cite{gu2009learning}. Particularly, the resulting method is named learning the shared subspace multi-task clustering (LSSMTC), which learns individual $k$-means clustering models simultaneously and a common shared subspace for each task. 
After that, many multi-task clustering methods have been proposed in the past few years 
\cite{gu2011learning,zhang2012multi,zhang2016multi,zhang2017multi}.
The abovementioned work has shown that MTC outperforms traditional single-task clustering algorithms in general. However, the existing MTC methods typically solve a non-convex problem and easily achieve a sub-optimal performance \cite{gu2009learning}.

In this paper, we adopt self-paced learning (SPL) paradigm to alleviate this problem and propose self-paced multi-task clustering (SPMTC). Concretely, self-paced learning is an example sampling strategy that is inspired by human learning process \cite{kumar2010self}. Instead of treating all the examples equally, SPL starts with ``easy'' examples and then gradually adds ``hard'' examples to train the model. Unlike curriculum learning \cite{bengio2009curriculum}, SPL does not need prior knowledge to determine the training order of the examples. The easiness of examples is defined by the off-the-shelf model itself \cite{Jiang2015SPCL}. It has been shown that SPL can avoid bad local optimum and can achieve a better generalization ability \cite{Jiang2015SPCL}. Traditional SPL model treats the selected samples equally. Recently, some variations of SPL are designed to not only choose examples but also assign weights to them \cite{Jiang2014easy,Pi2016Self,Ren2017Robust}.

Furthermore, outliers and noisy data which can negatively affect the clustering performance are generally common in multi-task clustering. To address this issue, a soft weighting strategy is designed in SPMTC. By assigning relatively small weights to noisy data or outliers, their negative influence will be significantly reduced. Overall, the main contributions of this paper are stated as below:
\begin{itemize}
\item We make use of SPL to address the non-convexity issue of multi-task learning in unsupervised setting.
To the best of our knowledge, this is the first work to apply self-paced learning to solve the multi-task clustering model.
\item The reconstruction error is used to assess how difficult it is to clustering a set of examples, with which a self-paced multi-task clustering (SPMTC) model is proposed. SPMTC helps obtain a better optimum and thus achieves better multi-task clustering performance.
\item A soft weighting strategy of SPL is employed to estimate the weights of data samples, according to which samples participate in training the MTC model. In this way, the negative influence of noisy data and outliers can be reduced and the clustering performance is further enhanced. An alternative strategy is developed to solve the proposed model and the convergence analysis is also given. 
\end{itemize}




\section{Related Work}
\label{sec:related}
\subsection{Multi-Task Clustering}
It has been shown that learning multiple related tasks simultaneously can be advantageous relative to learning these tasks independently \cite{evgeniou2004regularized,Liu2017Learning}. Multi-task learning (MTL) methods are firstly developed for classification problems and can be classified into two different types: regularization-based learning and joint feature learning. Regularization-based MTL minimizes the regularized objective function \cite{evgeniou2004regularized}, while joint feature learning methods capture the task relatedness by constraining all the tasks to a shared common feature set \cite{jalali2010dirty}.

To introduce multi-task learning in clustering, \cite{li2004document} proposed adaptive subspace iteration (ASI) that specifically identifies the subspace structure of multiple clusters, that is, projects all the examples into a subspace defined by the linear combination of the original feature space. Whereafter, \cite{gu2009learning} came up with the multi-task clustering method which combines the traditional $k$-means clustering and the ASI method by a balancing parameter, where the $k$-means clustering learns the individual clustering task and the ASI model learns the shared subspace of the multiple tasks. \cite{gu2011learning} proposed learning spectral kernel for multi-task clustering (LSKMTC) 
which learns a reproducing kernel hilbert space (RKHS) by formulating a unified kernel learning framework. 
After that, a number of multi-task clustering methods have been proposed \cite{zhang2012multi,zhang2017multi,xie2012multi,al2014multi,zhang2015convex,Zhang2016SAMTC}.

\subsection{Self-Paced Learning}
Curriculum learning (CL) is an instructor-driven model that organizes examples in a meaningful order under the premise of prior knowledge \cite{bengio2009curriculum}.  
Self-paced learning is a student-driven model that learns examples' easiness automatically \cite{kumar2010self}. Both CL and SPL order the examples by a certain rule rather than learning the model with them randomly. The difference between CL and SPL lies on the way they define the order of the examples: CL defines the order in advance by the prior knowledge while SPL defines the order by the loss computed by the model and updates the easiness according to the updated model \cite{Jiang2015SPCL}. \cite{Jiang2014easy} designed the soft weighting strategy to weaken the negative impact of noisy data. 
Due to its effectiveness, SPL has gained more and more attentions in various fields, e.g., computer vision \cite{Jiang2014easy,Tang2012Shifting,XU2018MSPL},  feature corruption \cite{Ren2017Balanced}, boosting learning \cite{Pi2016Self}, diagnosis of Alzheimer's Disease \cite{Que2017Regularized}, multi-class classification \cite{Ren2017Robust}, and so on. 


Recently, self-paced multi-task learning (SPMTL) has been proposed for supervised problems. For instance, \cite{Murugesan2017self} proposed a self-paced task selection method for multi-task learning, and \cite{li2017self} proposed a novel multi-task learning framework that learns the tasks by simultaneously considering the order of both tasks and instances. 
However, as we know, none of work has been done to enhance the multi-task clustering performance with self-paced learning. This paper fills this gap and proposes self-paced multi-task clustering in unsupervised setting. Soft weighting scenario for SPMTC is developed to further improve the clustering performance.

\section{Preliminaries}
Suppose $m$ clustering tasks are given, each associates with a set of data examples, $\mathcal{X}^{(k)}=\{\mathbf x^{(k)}_1, \mathbf x^{(k)}_2,\ldots, \mathbf x^{(k)}_{n_k}\}\in \mathbb{R}^{d}$, $k=1,2,\ldots,m$, where $d$ is the dimension of feature vectors for all tasks and $n_{k}$ is the number of examples in the $k$-th task. Let $\mathbf{X}^{(k)}=[\mathbf x^{(k)}_{1},\mathbf x^{(k)}_{2},\ldots,\mathbf x^{(k)}_{n_{k}}]\in \mathbb{R}^{d\times n_{k}}$ represent the $k$-th data set. Multi-task clustering seeks to group data examples of each task $\mathcal{X}^{(k)}$ into $c^{(k)}$ disjoint clusters. 
We assume $c^{(1)}=c^{(2)}=\dots=c^{(m)}=c$, which is generally assumed in the multi-task learning literature \cite{gu2009learning}.
Furthermore, we let the feature dimensionality $d$ be the same for all the tasks. In fact, we can use feature extraction or feature augmentation to make the dimensionality of all tasks the same. For example, the bag-of-words representation used in document analysis actually achieves this goal.

\subsection{Multi-Task Clustering}
Let us consider partitioning the $k$-th data set $\mathcal{X}^{(k)}$ into $c$ clusters by $k$-means, the following objective should be minimized:
\begin{equation}\label{eq:kmeans}
    \min_{\mathbf m_j^{(k)},1\le j\le c} \sum_{j=1}^c \sum_{\mathbf x_{i}^{(k)}\in C_{j}^{(k)}}||\mathbf x^{(k)}_{i}-\mathbf m^{(k)}_{j}||_{2}^{2},
\end{equation}
where $\mathbf m^{(k)}_{j}$ is the mean of cluster $C_{j}^{(k)}$, $C_{j}^{(k)}$ is the $j$-th cluster in the $k$-th task, and $\|\cdot\|_2$ means the $l_2$-norm of a vector. By letting $\mathbf M^{(k)}=[\mathbf m_{1}^{(k)},\mathbf m_{2}^{(k)},\ldots,\mathbf m_{c}^{(k)}] \in \mathbb{R}^{d\times c}$, equation (\ref{eq:kmeans}) can be rewritten as:
\begin{alignat}{2}
\setlength{\abovedisplayskip}{1pt}
\setlength{\belowdisplayskip}{1pt}
\min_{\mathbf M^{(k)},\mathbf P^{(k)}} &||\mathbf X^{(k)}-\mathbf M^{(k)}\mathbf P^{(k)\mathrm{T}}||_{F}^{2}\\
\mbox{s.t.}\quad  &\mathbf P^{(k)}\in\{0,1\}^{n_{k}\times c}. \nonumber
\end{alignat}
$\|\cdot\|_F$ denotes the Frobenius norm. $\mathbf P^{(k)}\in\{0,1\}^{n_{k}\times c}$ is a binary matrix that specifies each example into different clusters. It has the following assignment rules:
\begin{eqnarray}
\mathbf P_{ij}^{(k)}&= &~\left\{
    \begin{array}{ll}
      1,~~\text{if}~ \mathbf x_{i}^{(k)} \in C_{j}^{(k)},\\
      0,~~\text{otherwise}.\\
    \end{array}
  \right.    
\end{eqnarray}

Then, LSSMTC (learning the shared subspace multi-task clustering)  defines the objective of MTC as \cite{gu2009learning}:
\begin{align}\label{eq:MTC}
 \min_{\mathbf M,\mathbf M^{(k)}, \mathbf P^{(k)},\mathbf W} &\lambda\sum_{k=1}^{m}||\mathbf X^{(k)}-\mathbf M^{(k)}\mathbf P^{(k)\mathrm{T}}||_{F}^{2}\nonumber\\
             &+(1-\lambda)\sum_{k=1}^{m}||\mathbf W^{\mathrm{T}}\mathbf X^{(k)}-\mathbf M\mathbf P^{(k)\mathrm{T}}||_{F}^{2}\\
  \mbox{s.t.}\qquad\ &~\mathbf W^{\mathrm{T}}\mathbf W=\mathbf I,~\mathbf P^{(k)}  \geq 0,~k=1,\ldots,m,\nonumber
\end{align}
where $\mathbf W \in \mathbb{R}^{d\times l}$ is the orthonormal projection which learns a shared subspace across all the related tasks, and $\mathbf M=[\mathbf m_{1},\mathbf m_{2},\ldots,\mathbf m_{c}]\in \mathbb{R}^{l\times c}$ with $\mathbf m_{j}$ is the center of cluster $C_{j}$ of all tasks in the shared subspace. $\mathbf I$ is the identity matrix.  
The parameter $\lambda \in [0,1]$ controls the balance of the two parts in equation (\ref{eq:MTC}). 
The constraint of $\mathbf P$ is relaxed into non-negative continuous domain for optimization convenience.

\subsection{Self-Paced Learning} 
The self-paced learning strategy \cite{kumar2010self} iteratively learns the model parameter $\theta$ and the self-paced learning weight $\mathbf v$:
\begin{eqnarray}
\min_{\mathbf{\theta,v}}& \sum_{i=1}^{n} \mathbf v_{i}L(\mathbf x_{i}, \mathbf \theta)+f(\lambda,\mathbf v),
\end{eqnarray}
where $\lambda>0$, $\mathbf v\in[0,1]^n$ is the latent weighting variable, $n$ is the number of examples, and $L(\mathbf x_{i},\theta)$ is the corresponding loss function for example $\mathbf x_{i}$ in traditional classification or regression problems. $f(\lambda,\mathbf v)$ denotes the SPL regularizer. \cite{kumar2010self} defines $f(\lambda,\mathbf v)=-\lambda\sum_{i=1}^{n}\mathbf v_{i}$. 
When fix $\theta$, the optimal values of $\mathbf v$ is calculated by:
\begin{eqnarray}
 \mathbf v_{i}&= &~\left\{
    \begin{array}{ll}
      1,~~\text{if}~ L(\mathbf x_{i},\theta) \leq \lambda,\vspace{1ex}\\
      0,~~\text{otherwise},\\
    \end{array}
  \right.    
\end{eqnarray}
When $\lambda$ is small, a small number of examples are selected. As $\lambda$ grows gradually, more and more examples will be chosen to train the model until all the samples are chosen.

\section{Self-Paced Multi-Task Clustering}
\subsection{The Objective}
In this work we propose self-paced multi-task clustering (SPMTC) to address the non-convexity of multi-task clustering. By making use of self-paced learning paradigm which trains a model with examples from simplicity to complexity, SPMTC is able to avoid bad local optimum and find better solutions. When designing the SPL regularizer, SPMTC considers the easiness of examples not only from the example level, but also from the task level. Moreover, we develop a soft weighting technique of SPL to reduce the negative influence of noisy examples and outliers and to further enhance the multi-task clustering performance. 
The common optimization model of SPMTC is defined as:
\begin{eqnarray}\label{splmtc}
\min_{\mathbf M,\mathbf M^{(k)}, \mathbf P^{(k)}, \mathbf W, \mathbf{V}^{(k)}} \lambda_{1}\sum_{k=1}^{m}||(\mathbf X^{(k)}-\mathbf M^{(k)}\mathbf P^{(k)\mathrm{T}})\mathbf V^{(k)}||_{F}^{2}\nonumber\\
+(1-\lambda_{1})\sum_{k=1}^{m}||(\mathbf W^{\mathrm{T}}\mathbf X^{(k)}-\mathbf M\mathbf P^{(k)\mathrm{T}})\mathbf V^{(k)}||_{F}^{2}\\
-\sum_{k=1}^{m}f(\lambda_{2}^{(k)},\mathbf V^{(k)})~~~~~~~~~~~~~~~~~~~~~~~~~~~~~\nonumber  \\
  \mbox{s.t.}\ ~\mathbf W^{\mathrm{T}}\mathbf W=\mathbf I,~\mathbf P^{(k)}  \geq 0,~k=1,\ldots,m.\nonumber      
\end{eqnarray}

The first part of equation (\ref{splmtc}) contains $m$ independent $k$-means clustering tasks and is called \textit{within-task clustering}, while the second part involves clustering data of all tasks in the shared subspace and is referred as \textit{cross-task clustering}. $\lambda_{1}\in [0,1]$ is the trade-off parameter. The third part is the SPL regularizer. 
Matrix $\mathbf V^{(k)} = diag(\mathbf v^{(k)})$ where $\mathbf v^{(k)}= [\mathbf v^{(k)}_{1}, \mathbf v^{(k)}_{2},\ldots, \mathbf v^{(k)}_{n_{k}}] \in [0,1]^{n_k}$ denotes the weights of $n_{k}$ examples in the $k$-th task, $\lambda_{2}^{(k)}$ is the SPL controlling parameter for the $k$-th task, and $-f(\lambda_{2}^{(k)},\mathbf V^{(k)})$ is the corresponding SPL regularization term.

Traditional SPL regularizer assigns weights and selects examples from the entire data, which may suffer from a problem in multi-task setting. That is, if data from some tasks inherently have small loss values and thus own large weights, then these tasks will contribute more in the training process. By contract, those tasks whose data generally obtain small weights (which correspond to large loss values) participate little in the learning model. If this happens, the underlying relation among multiple tasks is not adequately utilized. In extreme cases, none or limited number of examples from these tasks are selected to train in the beginning of SPL.
Different from the traditional one, the developed SPL regularizer in this work is able to compute weights and choose examples from each task independently. This ensures that all the tasks have equal opportunity to participate in multi-task clustering process, and thus the relation among tasks can be sufficiently used.

\subsection{Optimization}
We firstly run LSSMTC a small number of iterations (which is set to 20 in the experiments) to obtain an initialization of model parameters $\mathbf M,\mathbf M^{(k)}, \mathbf P^{(k)}, \mathbf W$. By doing this, we can obtain an initialized estimation of reconstructing error for examples. Then, an alternative optimization method is designed to solve equation (\ref{splmtc}) and the optimization process mainly contains the following two steps.

\paragraph{Step 1: Fix model parameters, update $\mathbf V^{(k)}$.} For fixed MTC model parameter values (i.e., $\mathbf M,\mathbf M^{(k)}, \mathbf P^{(k)}, \mathbf W$), minimizing equation (\ref{splmtc}) is equivalent to solve
\begin{align}\label{eq:solve_V}
\min_{\mathbf v_{i}^{(k)}} &\lambda_{1}\sum_{k=1}^{m}\sum_{i=1}^{n_{k}} ||\mathbf x_{i}^{(k)}- [\mathbf M^{(k)}\mathbf P^{(k)\mathrm{T}}]_{ \centerdot i}||_{2}^{2}\mathbf v_{i}^{(k)}\nonumber\\
&+(1-\lambda_{1})\sum_{k=1}^{m}\sum_{i=1}^{n_{k}}||[\mathbf W \mathbf X^{(k)}- \mathbf M \mathbf P^{(k)\mathrm{T}}]_{\centerdot i}||_{2}^{2}\mathbf v_{i}^{(k)}\\
& -\sum_{k=1}^{m}f(\lambda_{2}^{(k)},\mathbf V^{(k)})\nonumber,
\end{align}
where  $[\mathbf M^{(k)}\mathbf P^{(k)\mathrm{T}}]_{ \centerdot i}$ means the $i$-th column of matrix $\mathbf M^{(k)}\mathbf P^{(k)\mathrm{T}}$.
Let $L(\mathbf x_{i}^{(k)})=\lambda_{1}||\mathbf x_{i}^{(k)}-\mathbf [\mathbf M^{(k)}\mathbf P^{(k)\mathrm{T}}]_{ \centerdot i}||_{2}^{2}+(1-\lambda_{1})||[\mathbf W \mathbf X^{(k)}- \mathbf M \mathbf P^{(k)\mathrm{T}}]_{\centerdot i}||_{2}^{2}$ be the reconstruction error of example $\mathbf x_{i}^{(k)}$, then equation (\ref{eq:solve_V}) becomes: 
\begin{equation}\label{eq:simple_V}
\min_{\mathbf v_{i}^{(k)}} \sum_{k=1}^{m}\sum_{i=1}^{n_{k}} L(\mathbf x_{i}^{(k)}) \mathbf v_{i}^{(k)}-\sum_{k=1}^{m}f(\lambda_{2}^{(k)},\mathbf V^{(k)}).
\end{equation}

In classification tasks, the labels of examples can be used to evaluate the loss values and assign weights. But, no supervised information is provided in clustering problems. Thus, we use the reconstruction error of an example as its loss value.
Intuitively, examples which are close to the centers in the original feature space and in the shared feature space obtain small reconstruction error, while examples that are far away from centers obtain large loss values and are considered as ``hard'' examples.
To solve equation (\ref{eq:simple_V}), we design two functions of $f(\lambda_{2}^{(k)},\mathbf V^{(k)})$, which correspond to hard weighting and soft weighting strategies, respectively.

\textbf{Hard weighting:} 
\begin{equation}\label{eq:hardweight}
    f(\lambda_{2}^{(k)},\mathbf{V}^{(k)})=\lambda_{2}^{(k)}\text{tr}{(\mathbf V^{(k)})}=\lambda_{2}^{(k)}||\mathbf{v}^{(k)}||_{1},
\end{equation}
where $\text{tr}(\cdot)$ denotes the trace of a matrix and $\|\cdot\|_1$ is the $l_1$-norm of a vector.
By substituting equation (\ref{eq:hardweight}) into equation (\ref{eq:simple_V}), we can easily find the optimal value of $\mathbf V^{(k)}$ given by:
\begin{eqnarray}\label{hard-weighting-v}
 \mathbf v_{i}^{(k)}&= &~\left\{
    \begin{array}{ll}
      1,~~\text{if}~ L(\mathbf x_{i}^{(k)}) \leq \lambda_{2}^{(k)},\vspace{1ex}\\
      0,~~\text{otherwise},\\
    \end{array}
  \right.    
\end{eqnarray}

\textbf{Soft weighting:} 
\begin{equation}\label{eq:mixweight}
    f(\lambda_{2}^{(k)},\mathbf V^{(k)})=\sum_{i=1}^{n_k} \gamma^{(k)} \ln(\mathbf v_i^{(k)}+\gamma^{(k)}/\lambda_{2}^{(k)}).
\end{equation}
Equation (\ref{eq:mixweight}) is referred as mixture soft weighting in \cite{Jiang2014easy,Ren2017Robust}. The SPL regularizer designed in this work is actually a variation of mixture soft weighting in multi-task setting.
By substituting equation (\ref{eq:mixweight}) into equation (\ref{eq:simple_V}), setting the derivation w.r.t. $\mathbf v_{i}^{(k)}$, and noting that $\mathbf v_{i}^{(k)}\in[0,1]$, the optimal $\mathbf V^{(k)}$ can be obtained by:
\begin{eqnarray}\label{mix-weighting-v}
 \mathbf v_{i}^{(k)}&= &~\left\{
    \begin{array}{ll}
      \quad\quad 1,\quad\quad\quad\text{if}~ L(\mathbf x_{i}^{(k)}) \leq \frac{\gamma^{(k)}\lambda_{2}^{(k)}}{\gamma^{(k)}+\lambda_{2}^{(k)}},\vspace{1ex}\\
      \quad\quad 0,\quad\quad\quad\text{if}~ L(\mathbf x_{i}^{(k)})\geq \lambda_{2}^{(k)},\vspace{2ex}\\
      {\gamma^{(k)}}/{L(\mathbf x_{i}^{(k)})} - {\gamma^{(k)}}/{\lambda_{2}^{(k)}},~~\text{otherwise}.
    \end{array}
  \right.    
\end{eqnarray}
For simplicity, parameter $\gamma^{(k)}$ is always set to $\frac{1}{2}\lambda_{2}^{(k)}$.

Both hard and soft weighting assign weights and select examples from each task independently. The difference is that hard weighting assign weight 1 to all the chosen examples, while soft weighting not only selects examples but also assigns weights (whose values range in $[0,1]$) to them. Thus, in soft weighting scenario, those ``hard'' examples (which are typically noisy data, outliers or overlapped data) generally obtain small weights and their negative influence is reduced.
\paragraph{Step 2: Fix $\mathbf V^{(k)}$, update model parameters.} For fixed $\mathbf V^{(k)},k=1,\ldots,m$, the last term of equation (\ref{splmtc}) is a constant value. Then, we update each of the four MTC model parameters (i.e., $\mathbf M,\mathbf M^{(k)}, \mathbf P^{(k)}, \text{~and~} \mathbf W$) when others are fixed.

\begin{itemize}
\item Update rule of $\mathbf M$: Optimizing (\ref{splmtc}) w.r.t. $\mathbf M$ while keeping  $\mathbf{W}$ and $\mathbf P^{(k)}$ fixed is equivalent to minimizing:
\begin{equation}
    J_{1}=\sum_{k=1}^{m}||(\mathbf W^{\mathrm{T}}\mathbf X^{(k)}-\mathbf M\mathbf P^{(k)\mathrm{T}})\mathbf V^{(k)}||_{F}^{2}.
\end{equation}
Representing $\mathbf X=[\mathbf X^{(1)},\mathbf X^{(2)},\ldots,\mathbf X^{(m)}]$, $\mathbf P=[\mathbf P^{(1)\mathrm{T}},\mathbf P^{(2)\mathrm{T}},\ldots,\mathbf P^{(m)\mathrm{T}}]^\mathrm{T}$ and\\ 
\begin{small}
$\mathbf V = 
 \begin{pmatrix} 
  \mathbf v_{1}^{(1)} & \cdots& 0    & \cdots & 0     & \cdots  & 0 \\
  \vdots&     \ddots&\vdots&\ddots  & \vdots& \ddots  & \vdots\\
  0 &\cdots&\mathbf v_{n_{1}}^{(1)}&\cdots&0       &\cdots &0 \\
  \vdots&     \ddots&\vdots&\ddots  & \vdots& \ddots  & \vdots\\
  0 &\cdots&0&\cdots&\mathbf v_{1}^{(m)}       &\cdots &0 \\
  \vdots&     \ddots&\vdots&\ddots  & \vdots& \ddots  & \vdots\\
  0 &\cdots&0&\cdots&0       &\cdots &\mathbf v_{n_{m}}^{(m)} \\
 \end{pmatrix}$, where $\mathbf{V}^{(k)}=
 \begin{pmatrix}
 \mathbf v_{1}^{(k)} & \cdots& 0\\
 \vdots&     \ddots&\vdots\\
 0 &\cdots&\mathbf v_{n_{k}}^{(k)}\\
 \end{pmatrix}$
\end{small} 
 , $J_{1}$ can be written as:
\begin{equation}
    J_{1}=||\mathbf{(W^{\mathrm{T}}X-MP^{\mathrm{T}})V}||_{F}^{2}.
\end{equation}
Setting $\frac{\partial J_{1}}{\partial \mathbf M}=\mathbf 0$, we have
\begin{equation}\label{update_m}
    \mathbf{M=W^{\mathrm{T}}XVV^{\mathrm{T}}P(P^{\mathrm{T}}VV^{\mathrm{T}}P)^{-1}}
\end{equation}

\item Update rule of $\mathbf M^{(k)}$: When fix $\mathbf P^{(k)}$, we can optimize equation (\ref{splmtc}) w.r.t. $\mathbf M^{(k)}$ by minimizing :
\begin{equation}
    J_{2}=||(\mathbf X^{(k)}-\mathbf M^{(k)}\mathbf P^{(k)\mathrm{T}})\mathbf V^{(k)}||_{F}^{2},
\end{equation}
setting $\frac{\partial J_{2}}{\partial \mathbf M^{(k)}}=\mathbf 0$, we can observe
\begin{equation}\label{update_mk}
    \mathbf M^{(k)}=\mathbf X^{(k)}\mathbf V^{(k)}\mathbf V^{(k)\mathrm{T}}\mathbf P^{(k)}(\mathbf P^{(k)\mathrm{T}}\mathbf V^{(k)}\mathbf V^{(k)\mathrm{T}}\mathbf P^{(k)})^{-1}.
\end{equation}
\item Update rule of $\mathbf P^{(k)}$: Solving (\ref{splmtc}) w.r.t. $\mathbf P^{(k)}$ while keeping $\mathbf M^{(k)}, \mathbf M$, and $\mathbf W$ fixed is equivalent to optimizing the following equation:
\begin{align}
J_{3}= &\lambda_{1}\sum_{k=1}^{m}||(\mathbf X^{(k)}-\mathbf M^{(k)}\mathbf P^{(k)\mathrm{T}})\mathbf V^{(k)}||_{F}^{2}\nonumber\\
               &+(1-\lambda_{1})\sum_{k=1}^{m}||(\mathbf W^{\mathrm{T}}\mathbf X^{(k)}-\mathbf M\mathbf P^{(k)\mathrm{T}})\mathbf V^{(k)}||_{F}^{2}\\
  \mbox{s.t.}\ &~\mathbf P^{(k)}  \geq 0,~k=1,\ldots,m.\nonumber
\end{align}
By introducing the Lagrangian multiplier $\boldsymbol{\alpha} \in \mathbb{R}^{n_{k}\times c}$, we obtain the Lagrangian function:
\begin{align}
L(\mathbf P^{(k)})= &\lambda_{1}\sum_{k=1}^{m}||(\mathbf X^{(k)}-\mathbf M^{(k)}\mathbf P^{(k)\mathrm{T}})\mathbf V^{(k)}||_{F}^{2}\nonumber\\
               &+(1-\lambda_{1})\sum_{k=1}^{m}||(\mathbf W^{\mathrm{T}}\mathbf X^{(k)}-\mathbf M\mathbf P^{(k)\mathrm{T}})\mathbf V^{(k)}||_{F}^{2}\\
               &-\text{tr}(\boldsymbol{\alpha} \mathbf P^{(k)\mathrm{T}}). \nonumber
\end{align}
Setting $\frac{\partial L(\mathbf P^{(k)})}{\partial \mathbf P^{(k)}}=\mathbf 0$, we observe
\begin{equation}
    \boldsymbol{\alpha} = -2\mathbf A+2\mathbf V^{(k)}\mathbf V^{(k)\mathrm{T}}\mathbf P^{(k)}\mathbf B,
\end{equation}
where $\mathbf A=\lambda_{1} \mathbf V^{(k)}\mathbf V^{(k)\mathrm{T}}\mathbf X^{(k)\mathrm{T}}\mathbf M^{(k)}+(1-\lambda_{1}) \mathbf V^{(k)} \mathbf V^{(k)\mathrm{T}} \mathbf X^{(k)}\mathbf W\mathbf M$, $\mathbf B=\lambda_{1}\mathbf M^{(k)}\mathbf M^{(k)\mathrm{T}}+(1-\lambda_{1})\mathbf{M^{\mathrm{T}}M}$. According to \cite{ding2010convex}, we can obtain the following update rule:
\begin{equation}\label{update_pk}
    \mathbf P_{ij}^{(k)}\leftarrow  \mathbf P_{ij}^{(k)}\sqrt{\frac{[\mathbf A^{+}+\mathbf V^{(k)}\mathbf V^{(k)\mathrm{T}}\mathbf P^{(k)} \mathbf B^{-}]_{ij}}{[\mathbf A^{-}+\mathbf V^{(k)}\mathbf V^{(k)\mathrm{T}}\mathbf P^{(k)} \mathbf B^{+}]_{ij}}}.
\end{equation}
Here, $\mathbf A=\mathbf A^{+}-\mathbf A^{-}$ and $\mathbf B=\mathbf B^{+}-\mathbf B^{-}$, where $\mathbf A_{ij}^{+}=\frac{|\mathbf A_{ij}|+\mathbf A_{ij}}{2}$ and $\mathbf A_{ij}^{-}=\frac{|\mathbf A_{ij}|-\mathbf A_{ij}}{2}$.

\item Update rule of $\mathbf W$: Optimizing (\ref{splmtc}) w.r.t. $\mathbf W$ while keeping $\mathbf M,\mathbf M^{(k)},\mathbf P^{(k)}$ fixed is equivalent to minimizing: 
\begin{align}
    J_{4}=&\sum_{k=1}^{m} ||(\mathbf W^{\mathrm{T}}\mathbf X^{(k)}-\mathbf M \mathbf P^{(k)\mathrm{T}})\mathbf V^{(k)}||_{F}^{2} \nonumber\\
         =&||(\mathbf W^{\mathrm{T}}\mathbf X -\mathbf M \mathbf P^{\mathrm{T}})\mathbf V||_{F}^{2}\\
      \mbox{s.t.}& \mathbf W^{\mathrm{T}}\mathbf W=\mathbf I. \nonumber
\end{align}
Substituting $\mathbf M=\mathbf{W^{\mathrm{T}}XVV^{\mathrm{T}}P(P^{\mathrm{T}}VV^{\mathrm{T}}P)^{-1}}$ into $J_{4}$, we have
\begin{align}
    J_{5}=&\text{tr}(\mathbf W^{\mathrm{T}}\mathbf X\mathbf V(\mathbf I -\mathbf V^{\mathrm{T}}\mathbf P(\mathbf P^{\mathrm{T}} \mathbf V\mathbf V^{\mathrm{T}}\mathbf P)\mathbf P^{\mathrm{T}}\mathbf V)\mathbf V^{\mathrm{T}}\mathbf X^{\mathrm{T}}\mathbf W),\nonumber\\
    \mbox{s.t.}&\mathbf W^{\mathrm{T}}\mathbf W=\mathbf I.
\end{align}
Then, the optimal $\mathbf W$ is composed of the eigenvectors of the following matrix corresponding to the $l$ smallest eigenvalues:
\begin{equation}\label{update_w}
    \mathbf X\mathbf V(\mathbf I -\mathbf V^{\mathrm{T}}\mathbf P(\mathbf P^{\mathrm{T}} \mathbf V\mathbf V^{\mathrm{T}}\mathbf P)\mathbf P^{\mathrm{T}}\mathbf V)\mathbf V^{\mathrm{T}}\mathbf X^{\mathrm{T}}.
\end{equation}
\end{itemize}

With fixed $\mathbf V^{(k)},k=1,\ldots,m$, step 2 actually solves a weighted version of the LSSMTC model \cite{gu2009learning}, whose convergence property has been guaranteed. Step 2 stops when converges or reaches the maximum number of iterations $T$, which is set to 50 in our experiments.

\begin{algorithm}[tbp] \small
\caption{Self-Paced Multi-Task Clustering. }
\label{alg:SPMTC}
\begin{algorithmic}
\REQUIRE Data sets $\mathcal{X}^{(k)},~k=1,2,\ldots,m$; \\
\qquad ~Maximum number of iterations $T$.
\ENSURE Partition matrices $\mathbf{P}^{(k)}$, $k=1,2,\ldots,m$.
\STATE Initialize model parameters  $\mathbf M,\mathbf M^{(k)}, \mathbf P^{(k)}, \mathbf W$.
\STATE Initialize SPL controlling parameters $\lambda_{2}^{(k)},~k=1,2,\ldots,m$.
\REPEAT
\STATE \textbf{Step 1} Fix $\mathbf{M,W},\mathbf M^{(k)}, \mathbf P^{(k)}$, update $\mathbf{V}^{(k)}$ by equation (\ref{hard-weighting-v}) or equation (\ref{mix-weighting-v}).
\STATE \textbf{Step 2} Fix $\mathbf{V}^{(k)}$, update $\mathbf{M,W},\mathbf M^{(k)}, \mathbf P^{(k)}$ as follows:
\REPEAT
\STATE Update $\mathbf{M}$ by equation (\ref{update_m}).
\STATE Update $\mathbf M^{(k)}$ by equation (\ref{update_mk}).
\STATE Update $\mathbf P^{(k)}$ by equation (\ref{update_pk}).
\STATE Udpate $\mathbf{W}$ by computing the eigenvectors of matrix (\ref{update_w}).
\UNTIL{convergence or $T$ is reached}
\STATE Update $\lambda_{2}^{(k)}$.
\UNTIL{all the examples are selected}
\RETURN $\mathbf{P}^{(k)}$, $k=1,2,\ldots,m$.
\end{algorithmic}
\end{algorithm}

SPMTC runs steps 1 and 2 iteratively. In the beginning, we initialize $\lambda_{2}^{(k)}~(k=1,\ldots,m)$ to let half of the data examples be chosen to train the model. Then, we increase $\lambda_2^{(k)}$ to add 10\% more examples from the $k$-th task in each of the following iterations.
SPMTC stops when all the data of all tasks are included. Then, the learned $\mathbf M,\mathbf M^{(k)}, \mathbf P^{(k)}, \text{~and~} \mathbf W$ are the final parameter values.
In the $k$-th task, the example $\mathbf x_i^{(k)}$ is assigned to the $j^*$-th cluster if $\mathbf P_{ij^*}^{(k)}>\mathbf P_{ij}^{(k)}$, $\forall j\neq j^*,j=1,2,\ldots,c$.
In summary, the pseudo code of SPMTC is described in Algorithm \ref{alg:SPMTC}.


\subsection{Algorithm Analysis}
The proposed SPMTC has the following properties: 
First, SPMTC inherits the advantage of avoiding bad local optima and thus can find better clustering solution.
Second, SPMTC with soft weighting assigns extremely small weight values to noises and outliers. Thus, these data samples will have extremely small influence in the training process.
This property is consistent with the loss-prior-embedding property of SPL described in \cite{Meng2017Theo}. That is, outliers and noises typically show large loss values and are associated with small weights, thus their negative affect can be significantly reduced. 
Third, the convergence of the proposed SPMTC model is theoretically guaranteed and the corresponding proof is given in Theorem \ref{theo:SPL_converge}.

\begin{theorem}
\label{theo:SPL_converge}
SPMTC is guaranteed to converge.
\begin{proof}
The proposed SPMTC iteratively updates model parameters ($\mathbf M,\mathbf M^{(k)}, \mathbf P^{(k)}, \mathbf W$) and self-paced learning weight $\mathbf V^{(k)}$.
In each iteration, when model parameters are updated, self-paced learning weight $\mathbf V^{(k)}$ can be obtained by a closed-form solution, i.e., equation (\ref{hard-weighting-v}) or equation (\ref{mix-weighting-v}).
When $\mathbf V^{(k)}$ is fixed, solving model parameters of  SPMTC is equivalent to solving a weighted version of the LSSMTC model, which has been theoretically prone to converge \cite{gu2009learning}. Specifically, solving model parameters of  SPMTC can be divided into four subproblems, which correspond to parameters $\mathbf M$, $\mathbf M^{(k)}$, $\mathbf P^{(k)}$, and $\mathbf W$, respectively. By alternatively updating each parameter with the other three fixed, the value of objective function (\ref{splmtc}) decreases correspondingly. Then, we obtain that the objective function value (\ref{splmtc}) is monotonically decreasing and is obviously lower bounded when its third part is a constant. Thus, SPMTC converges to a local minima when $\mathbf V^{(k)}$ is fixed.

As lambda grows and more data instances are selected for training, the model parameters and self-paced learning weight are updated accordingly. 
In the last iteration of SPMTC, when all the data instances are chosen, $\mathbf V^{(k)}$ is calculated and SPMTC finally converges.
\end{proof}
\end{theorem}

\section{Experiments}
\subsection{Experimental Setup}
\paragraph{\textbf{Data Sets}}  
We select two popular data sets based on which we design several binary clustering tasks since our focus is to verify the effectiveness of introducing self-paced learning to multi-task clustering. 
The 20 Newsgroups\footnote{http://people.csail.mit.edu/jrennie/20Newsgroups/} data set is consisted of roughly 20,000 newsgroup documents, which are partitioned into 20 different newsgroups, each corresponding to a different topic. Some of the newsgroups are very closely related to each other, while others are highly unrelated. In this paper, we use a subset of 20 Newsgroup which is also used in \cite{gu2009learning}, and it includes two different subsets: Comp vs Sci and Rec vs Talk.
Reuters-21578\footnote{http://www.cse.ust.hk/TL/} is a famous data base for text analysis. 
Among its five categories, orgs, people and places are the three biggest ones. We use three data sets orgs vs people, orgs vs places and people vs places generated by \cite{ling2008spectral}. {The WebKB dataset\footnote{http://www.cs.cmu.edu/afs/cs/project/theo-20/www/data/} collects webpages of computer science departments from different universities. It can be classified into 7 different categories: student, faculty, staff, department, course, project and other.} The detailed information of data sets can be found in Table \ref{data_details}.

\paragraph{\textbf{Comparing Methods}}
We compare the proposed SPMTC with the following single clustering methods: $k$-means (KM) \cite{MacQueen:some}, spectral clustering (SC) \cite{Luxburg:spectral}, and adaptive subspace iteration (ASI) \cite{li2004document}. We also use the above three methods to clustering all the tasks' data together and present the three methods: All KM, All SC, and All ASI, respectively. The multi-task clustering method LSSMTC \cite{gu2009learning} is also tested in the experiments. Our methods SPMTC-h and SPMTC-s denote SPMTC model with hard weighting and soft weighting, respectively.

\paragraph{\textbf{Parameter Setting}}
We always set the number of clusters equal to the true number of data labels. 
For LSSMTC, SPMTC-h and SPMTC-s, we tune $l$ and $\lambda_{1}$ in the same way and report the best results, where $l\in \{2, 4, 8, 16\}$, and $\lambda_{1}\in\{0.05,0.1,0.15,0.2,\ldots,0.95\}$. 

\begin{table} 
\centering
\caption{Data sets used in the experiments.}
\label{data_details}
\begin{tabular}{l|c|c|c|c}
\hline
\text{\small Data Set}    & \text{\small TaskID} & \text{\small\#Sample} & \text{\small\#Feature} & \text{\small\#Class} \\
\hline
\multirowcell{2}{\text{\small COMPvsSCI}} 
& Task 1 & 1875 & 2000 &2\\
& Task 2 & 1827 & 2000 &2\\\hline 
\multirowcell{2}{\text{\small RECvsTALK}} 
& Task 1 & 1844 & 2000  &2\\
& Task 2 & 1545 & 2000  &2\\\hline
\multirowcell{2}{\text{\small ORGSvsPEOPLE}} 
& Task 1 & 1237 & 4771  &2\\
& Task 2 & 1208 & 4771  &2\\\hline
\multirowcell{2}{\text{\small ORGSvsPLACES}} 
& Task 1 & 1016 & 4405  &2\\
& Task 2 & 1043 & 4405  &2\\\hline 
\multirowcell{2}{\text{\small PEOPLEvsPLACES}} 
& Task 1 & 1077 & 4562  &2\\
& Task 2 & 1077 & 4562  &2\\\hline                             \multirowcell{4}{\text{WebKB}} 
& Task 1 & 226 & 2500 &6\\
& Task 2 & 252 & 2500 &6\\
& Task 3 & 255 & 2500 &6\\
& Task 4 & 307 & 2500 &6\\
\hline              
\end{tabular}  
\end{table}

\begin{table}[!h]
  \centering
  \caption{Results on COMPvsSCI.} \label{comp_sci}
    \begin{tabular}{l|c|c|c|c}
    \hline
    \multicolumn{1}{c|}{\multirow{2}[4]{*}{Methods}} & \multicolumn{2}{c|}{Task1} & \multicolumn{2}{c}{Task2} \\
\cline{2-5}          & \multicolumn{1}{c|}{ACC} & \multicolumn{1}{c|}{NMI} & \multicolumn{1}{c|}{ACC} & \multicolumn{1}{c}{NMI} \\
    \hline
    KM & \multicolumn{1}{c|}{69.50} & \multicolumn{1}{c|}{26.21} & \multicolumn{1}{c|}{58.22} & \multicolumn{1}{c}{7.30} \\
    SC & \multicolumn{1}{c|}{50.88} & \multicolumn{1}{c|}{6.66} & \multicolumn{1}{c|}{51.94} & \multicolumn{1}{c}{4.12} \\
    ASI   & \multicolumn{1}{c|}{87.86} & \multicolumn{1}{c|}{52.33} & \multicolumn{1}{c|}{69.44} & \multicolumn{1}{c}{18.93} \\
    All KM & \multicolumn{1}{c|}{66.89} & \multicolumn{1}{c|}{18.75} & \multicolumn{1}{c|}{53.72} & \multicolumn{1}{c}{2.04} \\
    All SC & \multicolumn{1}{c|}{51.57} & \multicolumn{1}{c|}{0.00} & \multicolumn{1}{c|}{51.94} & \multicolumn{1}{c}{4.12} \\
    All ASI & \multicolumn{1}{c|}{87.76} & \multicolumn{1}{c|}{52.09} & \multicolumn{1}{c|}{\textbf{84.21}} & \multicolumn{1}{c}{\textbf{37.48}} \\
    LSSMTC & \multicolumn{1}{c|}{89.92} & \multicolumn{1}{c|}{56.91} & \multicolumn{1}{c|}{75.40} & \multicolumn{1}{c}{29.64} \\
    SPMTC-h & \multicolumn{1}{c|}{90.87} & \multicolumn{1}{c|}{59.44} & \multicolumn{1}{c|}{81.72} & \multicolumn{1}{c}{35.05} \\
    SPMTC-s & \multicolumn{1}{c|}{\textbf{92.02}} & \multicolumn{1}{c|}{\textbf{61.34}} & \multicolumn{1}{c|}{76.31} & \multicolumn{1}{c}{29.20}\\
    \hline
    \end{tabular}%
\end{table}%

\begin{table}[!h]
  \centering
  \caption{Results on RECvsTALK.}\label{rec_talk}
    \begin{tabular}{l|c|c|c|c}
    \hline
    \multicolumn{1}{c|}{\multirow{2}[4]{*}{Methods}} & \multicolumn{2}{c|}{Task1} & \multicolumn{2}{c}{Task2} \\
\cline{2-5}          & ACC   & NMI   & ACC   & NMI \\
    \hline
    KM & 63.72& 12.57& 58.33& 3.92\\
    SC & 57.66& 4.87& 58.87 & 4.20\\
    ASI   & 67.02& 10.80& 63.14 & 5.02\\
    All KM & 64.96& 12.44& 59.20& 3.34\\
    All SC & 56.18& 3.33& 55.47& 0.61\\
    All ASI & 67.29& 12.29& 64.23& 8.69\\
    LSSMTC & 91.85 & 61.23& 84.16& 44.52\\
    SPMTC-h & 92.00& 61.47& 84.76& 45.47\\
    SPMTC-s & \textbf{92.56}& \textbf{64.05}& \textbf{88.69}& \textbf{52.61}\\
    \hline
    \end{tabular}%
  \label{tab:addlabel}%
\end{table}%

\begin{table}[!t]
  \centering
  \caption{Results on ORGSvsPEOPLE.}\label{orgs_people}
    \begin{tabular}{l|c|c|c|c}
    \hline
    \multicolumn{1}{c|}{\multirow{2}[4]{*}{Methods}} & \multicolumn{2}{c|}{Task1} & \multicolumn{2}{c}{Task2} \\
\cline{2-5}          & ACC   & NMI   & ACC   & NMI \\
    \hline
    KM& 54.62& 0.62& 52.08& 0.12\\
    SC & 63.86& 11.34& 52.81& 1.48\\
    ASI   & 61.35& 4.52& 59.67 & 3.56\\
    All KM & 54.66& 0.49& 52.67& 0.19\\
    All SC & 52.06& 0.73& 52.15& 0.33\\
    All ASI & 62.37& 5.15& 62.30& 5.01\\
    LSSMTC & 71.45& 14.05& 60.26 & 4.08\\
    SPMTC-h & 73.27& 16.18& 62.12& 5.14\\
    SPMTC-s &\textbf{73.67} &\textbf{17.31}&\textbf{68.62} &\textbf{12.33}\\
    \hline
    \end{tabular}%
\end{table}%

\begin{table}[!h]
  \centering
  \caption{Results on ORGSvsPLACES.}\label{orgs_places}
    \begin{tabular}{l|c|c|c|c}
    \hline
    \multicolumn{1}{c|}{\multirow{2}[4]{*}{Methods}} & \multicolumn{2}{c|}{Task1} & \multicolumn{2}{c}{Task2}  \\
\cline{2-5}          & ACC   & NMI   & ACC   & NMI  \\
    \hline
    KM & 55.17& 0.73& 58.39& 1.79 \\
    SC & 65.19 & 9.63 & 61.12 & 3.66\\
    ASI   & \textbf{67.26} & {9.79}& 62.91& 5.43\\
    All KM & 53.25& 0.71& 59.18 & 1.89\\
    All SC & 57.57& 0.02& 60.59& 3.01\\
    All ASI & \textbf{66.71}& {9.08}& 64.56& 6.91\\
    LSSMTC &\textbf{67.11} & \textbf{10.63} & 62.40& 5.65\\
    SPMTC-h &\textbf{68.09}& \textbf{11.10}& 68.73& 9.69\\
    SPMTC-s &\textbf{67.24}& \textbf{10.03}&\textbf{71.14}& \textbf{12.45}\\
    \hline
    \end{tabular}%
\end{table}%

\paragraph{\textbf{Evaluation Measures}}
The accuracy (ACC) and the normalized mutual information metric (NMI) are adopted in this paper to evaluate the performance of comparing methods. 
Larger values of ACC or NMI indicate better clustering performance.
The average results of 20 independent runs of each method are reported. 
$t$-test is used to assess the statistical significance of the results at 5\% significance level.

\subsection{Results on Real Data Sets}
The clustering results are shown in Tables \ref{comp_sci}-\ref{WebKB}. In each column, we highlight the best and comparable results. First, we can observe from these tables that the multi-task clustering methods (i.e., ASI, LSSMTC, SPMTC-h, and SPMTC-s) generally outperform the single-task clustering methods. Second, compared with LSSMTC, higher ACC and NMI values are achieved by SPMTC-h in most of time. This demonstrates the advantage of applying SPL in the MTC model.
Third, All ASI performs the best for task 2 of COMPvsSCI and task 1 of WebKB, and SC outperforms other methods for task 2 of PEOPLEvsPLACES. Besides, the proposed SPMTC-s always obtains the best or comparable performance, indicating the usefulness of evaluating and utilizing weights of data instances in the model according to their loss values.

\begin{table}[!t]
  \centering
  \caption{Results on PEOPLEvsPLACES.}\label{people_places}
    \begin{tabular}{l|c|c|c|c}
    \hline
    \multicolumn{1}{c|}{\multirow{2}[4]{*}{Methods}} & \multicolumn{2}{c|}{Task1} & \multicolumn{2}{c}{Task2} \\
\cline{2-5}          & ACC   & NMI   & ACC   & NMI \\
    \hline
    KM & 58.18& 0.17& 61.69& 3.09 \\
    SC & 61.16&0.91&\textbf{63.53} & \textbf{6.31}\\
    ASI   & 62.49& 2.99 &56.19& 0.46\\
    All KM & 58.25& 0.11& 61.61& 3.09\\
    All SC & 59.91& 0.62& 60.74& 3.83\\
    All ASI & 62.76& 3.28& 61.10 & 2.67\\
    LSSMTC & 62.49 & 4.77& 55.51& 0.87\\
    SPMTC-h & 63.41 & 5.55& 55.98& 0.80\\
    SPMTC-s & \textbf{67.43} & \textbf{10.92}& 57.31& 1.03 \\
    \hline
    \end{tabular}%
\end{table}%

\begin{table}[!t]
  \centering
  \caption{Results on WebKB.}\label{WebKB}
    \begin{tabular}{c|r|r|r|r|r|r|r|r}
    \hline
    \multirow{2}[4]{*}{Methods} & \multicolumn{2}{c|}{Task1} & \multicolumn{2}{c|}{Task2} & \multicolumn{2}{c|}{Task3} & \multicolumn{2}{c}{Task4} \\
  \cline{2-9}        & \multicolumn{1}{c|}{ACC} & \multicolumn{1}{c|}{NMI} & \multicolumn{1}{c|}{ACC} & \multicolumn{1}{c|}{NMI} & \multicolumn{1}{c|}{ACC} & \multicolumn{1}{c|}{NMI} & \multicolumn{1}{c|}{ACC} & \multicolumn{1}{c}{NMI} \\
    \hline
    KM    & 59.86 & 13.71 & 55.71 & 12.40 & 54.19 & 10.93 & 55.60 & 14.21 \\
    SC    & 44.95 & 20.34 & 50.11 & 16.17 & 46.43 & 24.53 & 54.03 & 29.23 \\
    ASI   & {63.53} & 26.27 & 63.65 & 26.75 & 58.43 & 23.88 & 62.86 & 30.55 \\
    All KM & 61.15 & 14.75 & 61.42 & 12.40 & 51.29 & 5.51 & 57.85 & 12.48 \\
    All SC & 61.76 & 23.92 & 53.01 & 19.50 & {60.31} & 18.81 & 66.12 & \textbf{37.97} \\
    All ASI & \textbf{66.72} & \textbf{35.09} & 64.92 & 23.74 & {60.62} & 24.31 & 62.93 & 33.02 \\
    LSSMTC & 63.09 & 24.82 & 60.00 & {25.61} & \textbf{61.09} & {21.31} & 58.89 & 26.47 \\
    SPMTC-h & \textbf{64.34} & 23.80 & 64.29 & 26.00 & 57.18 & 17.61 & 65.60 & 29.24 \\
    SPMTC-m & \textbf{64.07} & 28.84 & \textbf{66.19} & \textbf{27.79} & \textbf{62.20} & \textbf{26.99} & \textbf{69.51} & \textbf{39.64} \\
    \hline
    \end{tabular}%
\end{table}%

\begin{figure}[!t]
\centering
\subfigure[ACC]{\includegraphics[width=0.49\columnwidth]{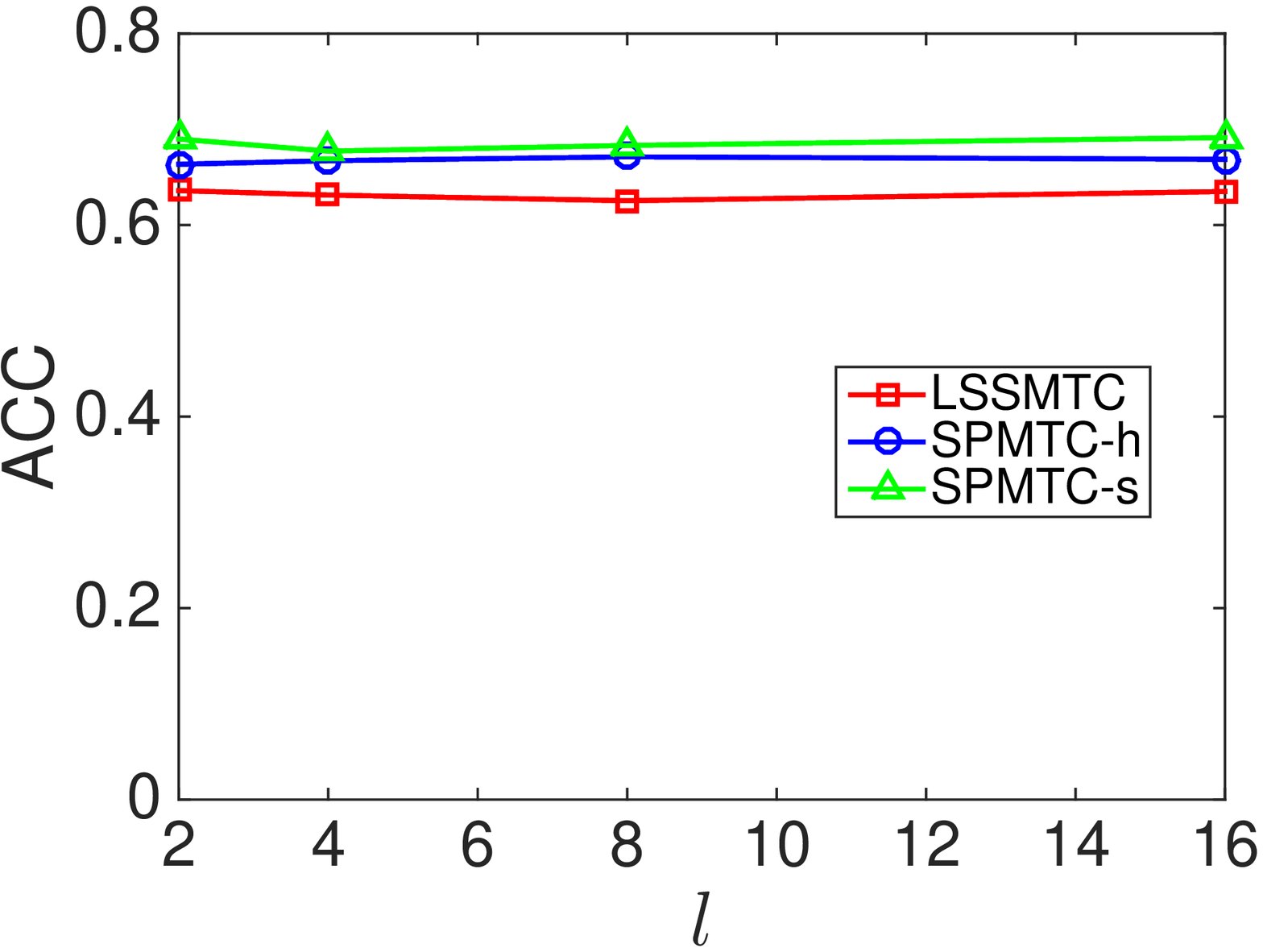}}
\subfigure[NMI]{\includegraphics[width=0.49\columnwidth]{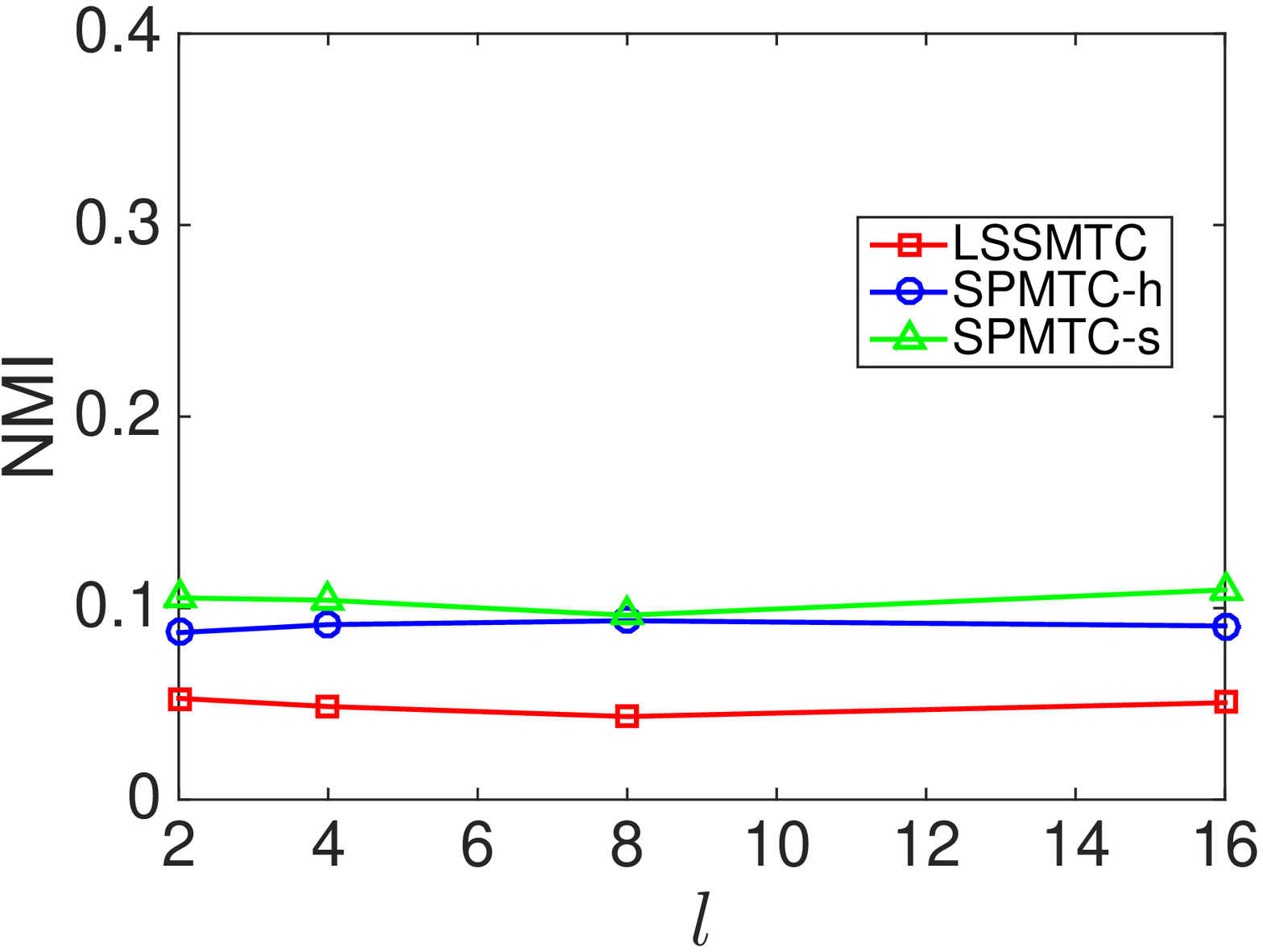}}
\vspace{-10pt}
\caption{Sensitivity analysis on PEOPLEvsPLACES.}
\label{fig:sensitivity}
\end{figure}

\subsection{Sensitivity and Time Complexity Analysis}
In real multi-task clustering applications, if the tasks are more related, the \textit{cross-task clustering} term should be considered more and $\lambda$ should be set to a smaller value, and vice versa.
In this section, we set $\lambda=0.5$ and test the sensitivity of SPMTC w.r.t. the feature dimension of the shared space $l$. 
Fig.~\ref{fig:sensitivity} shows the results of LSSMTC, SPMTC-h, and SPMTC-s on PEOPLEvsPLACES. We can see that different choices of $l$ do not significantly affect the performance of SPMTC.

SPMTC needs to solve a weighted version of LSSMTC several times (which is much less than 10 in the experiments). But, SPL generally speeds the convergence because each iteration of SPL takes the model parameter values trained by the previous iteration as initializations.
Thus, the time complexity of the proposed SPMTC is similar with LSSMTC. For instance,  the average running time of LSSMTC and SPMTC-h on PEOPLEvsPLACES is 3.1 and 4.2 seconds, respectively.



\section{Conclusion and Future Work}
\label{sec:Conclusions}
We have proposed self-paced multi-task clustering (SPMTC)  to alleviate the non-convexity issue of traditional multi-task clustering. Furthermore, we develop a soft weighting scheme of self-paced learning for SPMTC model to further enhance its clustering performance. 
The convergence analysis of SPMTC is given and its effectiveness is demonstrated by experimental results on real data sets.
We are interested in extending the framework proposed in this paper to more general situations in our future work, e.g., the number of clusters or feature dimensionality is different among different tasks. 


\section*{Acknowledgments}
This paper was in part supported by Grants from the Natural Science Foundation of China (Nos. 61806043, 61572111, and 61806043), a Project funded by China Postdoctoral Science Foundation (No. 2016M602674), a 985 Project of UESTC (No. A1098531023601041), and two Fundamental Research Funds for the Central Universities of China (Nos. ZYGX2016J078 and ZYGX2016Z003).


\bibliographystyle{elsarticle-num} 
\bibliography{Yazhou}

\end{document}